\documentclass[conference]{IEEEtran}
\IEEEoverridecommandlockouts

\usepackage[T1]{fontenc}
\usepackage{cite}
\usepackage{amsmath,amssymb,amsfonts}
\usepackage{algorithmic}
\usepackage{graphicx}
\usepackage{stfloats}
\usepackage{textcomp}
\usepackage{xcolor}
\usepackage{booktabs}
\usepackage{url}
\usepackage{xspace}
\usepackage{placeins}
\usepackage{multirow}

\def\BibTeX{{\rm B\kern-.05em{\sc i\kern-.025em b}\kern-.1667em\lower.7ex\hbox{E}\kern-.125emX}}
\newcommand{\ours}{QSRadar}

\begin{document}

\title{QSRadar: Query-Conditioned Sparse Radar--Camera Fusion for Efficient 3D Object Detection}

\author{%
\IEEEauthorblockN{Chenhao Yan\IEEEauthorrefmark{1}\IEEEauthorrefmark{2},
Hongyu Liu\IEEEauthorrefmark{1}\IEEEauthorrefmark{2},
Xiang Cai\IEEEauthorrefmark{1}\IEEEauthorrefmark{2}\IEEEauthorrefmark{3},
Yifei Sun\IEEEauthorrefmark{1}, and
Zhe Zhang\IEEEauthorrefmark{1}\IEEEauthorrefmark{2}\IEEEauthorrefmark{4}}
\IEEEauthorblockA{\IEEEauthorrefmark{1}School of Electronic, Electrical and Communication Engineering,\\
University of Chinese Academy of Sciences, Beijing, China}
\IEEEauthorblockA{\IEEEauthorrefmark{2}National Key Laboratory of Microwave Imaging,\\
Aerospace Information Research Institute, Chinese Academy of Sciences, Beijing, China}
\IEEEauthorblockA{\IEEEauthorrefmark{3}Suzhou Key Laboratory of Microwave Imaging, Processing and Application Technology,\\
Suzhou Aerospace Information Research Institute, Suzhou, China}
\IEEEauthorblockA{\IEEEauthorrefmark{4}School of Electronics Engineering,\\
Aerospace Information Technology University, Jinan, Shandong, China}
\IEEEauthorblockA{E-mail: \{yanchenhao22, liuhongyu226, caixiang26, sunyifei233\}@mails.ucas.ac.cn;\\
zhangzhe01@aircas.ac.cn}}

\maketitle
\suppressfloats[t]

\begin{abstract}
Reliable 3D object detection for autonomous trucks requires both the semantic detail provided by cameras and the geometric and motion cues measured by 4D radar. Existing radar--camera detectors commonly construct global radar bird's-eye-view (BEV) representations, whose computational and memory costs grow with the perception range. Sparse query-based detectors offer greater efficiency, but global feature construction can obscure point-level radar geometry and motion information. We present \ours{}, a query-conditioned sparse radar fusion framework for efficient long-range 3D detection. Built on a Sparse4Dv3-style visual--temporal detector, \ours{} retrieves a compact set of raw radar points around each evolving object hypothesis. The selected points are represented in an anchor-aligned coordinate frame, augmented with motion residuals, radar cross section, and sweep timing, and aggregated by a query-conditioned set encoder. Radar neighborhoods are retrieved again after each anchor refinement, coupling evidence selection with object estimation. Under the official MAN TruckScenes validation protocol, \ours{} achieves 38.09 mAP and 52.20 NDS, compared with 37.08 mAP and 50.78 NDS for our reproduced Sparse4D-Radar global Radar-BEV baseline. In matched tests on an RTX 4090 D, \ours{} reduces latency by 45\%, peak memory usage by 49\%, and GFLOPs by 43\%.
\end{abstract}

\begin{figure}[t]
\centering{}
\includegraphics[width=\columnwidth]{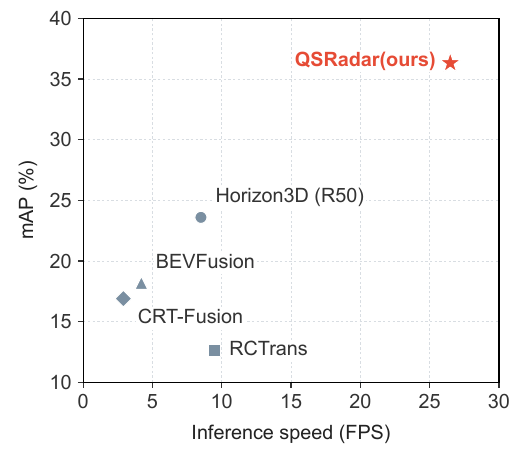}
\caption{mAP--FPS comparison on the MAN TruckScenes validation set. Results at 256$\times$864 for RCTrans, CRT-Fusion, BEVFusion, and Horizon3D are compiled from the public Horizon3D report~\cite{horizon3d}; QSRadar's FPS at this resolution is measured on an RTX 4090 D. Hardware and timing scopes differ across methods, so this plot provides efficiency context rather than a strictly controlled speed ranking.}
\label{fig:sota_fps_map}
\end{figure}


\section{Introduction}
Accurate 3D object detection is essential for autonomous driving: a vehicle must estimate the positions, dimensions, orientations, and motion of surrounding objects before planning safe actions. These demands are particularly challenging for autonomous trucks. Compared with passenger vehicles, trucks typically have higher sensor mounting positions and larger self-occluding structures, such as trailers, and operate in both long-range highway traffic and complex terminal environments. Reliable perception over an extended spatial range is therefore a fundamental challenge, not merely a straightforward extension of passenger-vehicle perception.

Cameras and 4D millimeter-wave radar provide complementary observations of driving scenes. Multi-view cameras offer rich appearance and semantic cues, but depth estimation is ambiguous, and visibility can deteriorate under darkness, rain, fog, and glare. In contrast, 4D radar measures metric range, elevation, and motion-related information and is relatively robust to adverse weather~\cite{radar_survey,kradar}. Radar returns, however, are often sparse and noisy and may not support fine-grained object recognition on their own. Radar--camera fusion thus combines radar's geometric and motion cues with the rich semantic information provided by vision.

Recent radar--camera detectors broadly follow two representation paradigms. BEV-based methods align modalities on a shared spatial grid and provide useful scene-level context, but their computational and memory costs increase with spatial extent and resolution~\cite{bevformer,bevfusion,rcfusion,rcbevdet}. Query-based methods instead maintain a compact set of object hypotheses, making them attractive for long-range perception and onboard inference~\cite{racformer,sparc}. Nevertheless, when radar points are compressed into a global BEV feature map, or sampled without accounting for evolving object hypotheses, sparse returns may be diluted by irrelevant regions and their motion attributes may be underused. This creates a tension between efficient fusion and effective access to raw radar evidence.

We study this problem on MAN TruckScenes, a multimodal benchmark for autonomous trucking with multi-view cameras, multiple 4D radars, long-range annotations, and truck-specific driving conditions~\cite{mantruckscenes}. Our central observation is that the current 3D anchor naturally provides a condition for selecting object-relevant radar evidence. Instead of constructing a global Radar-BEV branch, we let each query retrieve a compact local neighborhood from accumulated raw radar points. We then express these points in the query box coordinate frame and explicitly encode position, velocity residuals, signal strength, and time lag. This design retains object-level geometric and motion cues without the cost of dense radar feature construction.

Our framework builds on the sparse visual--temporal query architecture of Sparse4Dv3~\cite{sparse4dv3}. \ours comprises three coupled operations: size-adaptive local radar retrieval, hypothesis-aligned geometric and motion encoding, and query-conditioned set aggregation. Local radar features are fused with visual features at every decoder layer, and neighborhoods are retrieved again after anchor updates. Radar evidence is therefore realigned as the 3D hypotheses evolve. Using the official evaluator, we compare against camera-only and global BEV baselines in terms of detection accuracy, range-dependent performance, and inference efficiency. Fig.~\ref{fig:sota_fps_map} places our detector in the context of publicly reported MAN TruckScenes accuracy and throughput results.

Our main contributions are as follows:
\begin{itemize}
    \item We propose \ours{}, an efficient query-conditioned radar--camera fusion framework that accesses local raw radar evidence directly, replacing dense global Radar-BEV feature extraction. Layer-wise anchor refinement, neighborhood retrieval, and fusion form an iterative loop that couples radar evidence selection with object estimation.
    \item We introduce an anchor-aligned local radar encoder that combines top-$K$ retrieval with a size-dependent radius constraint, box-frame geometry and motion residuals, query-conditioned channel gating, and masked set aggregation. It encodes variable-size, potentially empty local radar observations into object-relevant features.
    \item We evaluate \ours{}under the official MAN TruckScenes protocol using camera-only and global Radar-BEV fusion baselines. Comparisons of detection accuracy, range-dependent performance, and computational efficiency demonstrate competitive accuracy with substantial reductions in inference latency, computation, and memory usage.
\end{itemize}

\section{Related Work}
\subsection{Radar--Camera 3D Object Detection}
Early radar--camera detectors associate radar measurements with image detections or proposals to provide depth and range cues~\cite{centerfusion,craft}. Later methods project both modalities into BEV representations and use radar-guided view transformation or cross-modal feature alignment~\cite{bevfusion,crn,rcfusion,rcbevdet}. Although these designs improve geometric consistency, they commonly rely on dense spatial features and global feature interactions.

Another line of work uses sparse object queries to reduce the cost of multimodal interactions. RaCFormer introduces query-based radar--camera fusion, while SpaRC explores sparse radar--camera feature interactions for 3D detection~\cite{racformer,sparc}. These methods are closely related to our setting because they represent scenes with compact sets of object hypotheses. Our approach differs in directly retrieving multi-sweep raw radar points around evolving queries while retaining point-level motion and timing attributes throughout aggregation.

Recent studies also investigate radar feature generation, semantic--geometric fusion, evidence association, and motion-aware radar processing~\cite{hgsfusion,sgdet3d,sgdet3dpp,drnet}. These efforts highlight the importance of radar representation quality and the association between measurements and object hypotheses. We complement them from an efficiency perspective: each query retrieves local raw points, encodes them in a query-aligned frame, and aggregates them using a lightweight set encoder at every decoder refinement stage.

\subsection{Sparse Queries and Temporal 3D Detection}
Sparse4D and its successors represent objects using compact sets of 3D anchors and instance features, with deformable sampling used to gather multi-view image evidence~\cite{sparse4d,sparse4dv3}. An instance bank propagates historical object queries after ego-motion alignment, enabling temporal modeling within the sparse representation. We retain this visual--temporal backbone and focus on injecting raw radar evidence into the same query representation. Rather than replacing the visual decoder or introducing a separate temporal memory, we add a query-local radar path compatible with the inherited sparse decoder.

\subsection{Long-Range Perception for Autonomous Trucks}
Long-range detection is challenging because objects occupy fewer image pixels, yield fewer radar returns, and exhibit larger inter-frame displacements. MAN TruckScenes exposes these challenges in autonomous trucking, including long-range annotations, trailer occlusion, complex terminal scenes, and diverse environmental conditions~\cite{mantruckscenes}. Recent methods such as Horizon3D and EviScope further highlight the importance of sparse representations and evidence-aware sampling at trucking-scale ranges~\cite{horizon3d,eviscope}. We likewise target efficiency, but select raw radar points conditioned on each object query and preserve their geometric and motion attributes in a query-aligned coordinate frame.

\section{Method}
\subsection{Problem Formulation and Motivation}
Given synchronized multi-view images and multi-sweep radar measurements at time $t$, our goal is to predict a set of 3D objects in the current reference frame. Let the transformed radar points be
\begin{equation}
    \mathcal{R}_t=\{r_j\}_{j=1}^{N},\qquad
    r_j=[x_j,y_j,z_j,v^x_j,v^y_j,s_j,\Delta t_j].
\end{equation}
where $(x_j,y_j,z_j)$ denotes position in the current reference frame, $(v^x_j,v^y_j)$ are the horizontal components of the ego-motion-compensated radial velocity vector expressed in that frame, and $s_j$ is radar cross section (RCS). The nonnegative time lag $\Delta t_j=t_{\mathrm{latest}}^{(k(j))}-t_j\geq0$ is measured relative to the latest sweep from the same radar sensor, where $k(j)$ identifies the sensor associated with point $j$ and $t_j$ is its sweep timestamp. The detector outputs 3D boxes, class labels, and confidence scores.

We adopt a Sparse4Dv3-style visual--temporal query detector as the backbone. At decoder layer $l$, the detector maintains an instance feature $f_i^{(l)}$ and a 3D anchor $a_i^{(l)}$ for each query. The anchor contains the object center, dimensions, orientation, and horizontal velocity. Visual features and historical queries establish and progressively update these hypotheses, while the local radar branch supplies additional evidence conditioned on the current anchors.

A common radar--camera pipeline first voxelizes the radar scene and encodes it into global Radar-BEV features, which are then sampled near the current anchors~\cite{sparse4dradar,rcfusion,rcbevdet}. This introduces a mismatch in representation: downstream queries directly access only a small set of anchor-conditioned locations, yet voxelization and the BEV backbone still process large regions not directly sampled by the current hypotheses. Motivated by this observation, we use the location prior supplied by each anchor to retrieve nearby raw radar points directly, without constructing a global Radar-BEV representation.

\begin{figure*}[t]
\centering
\includegraphics[width=\textwidth]{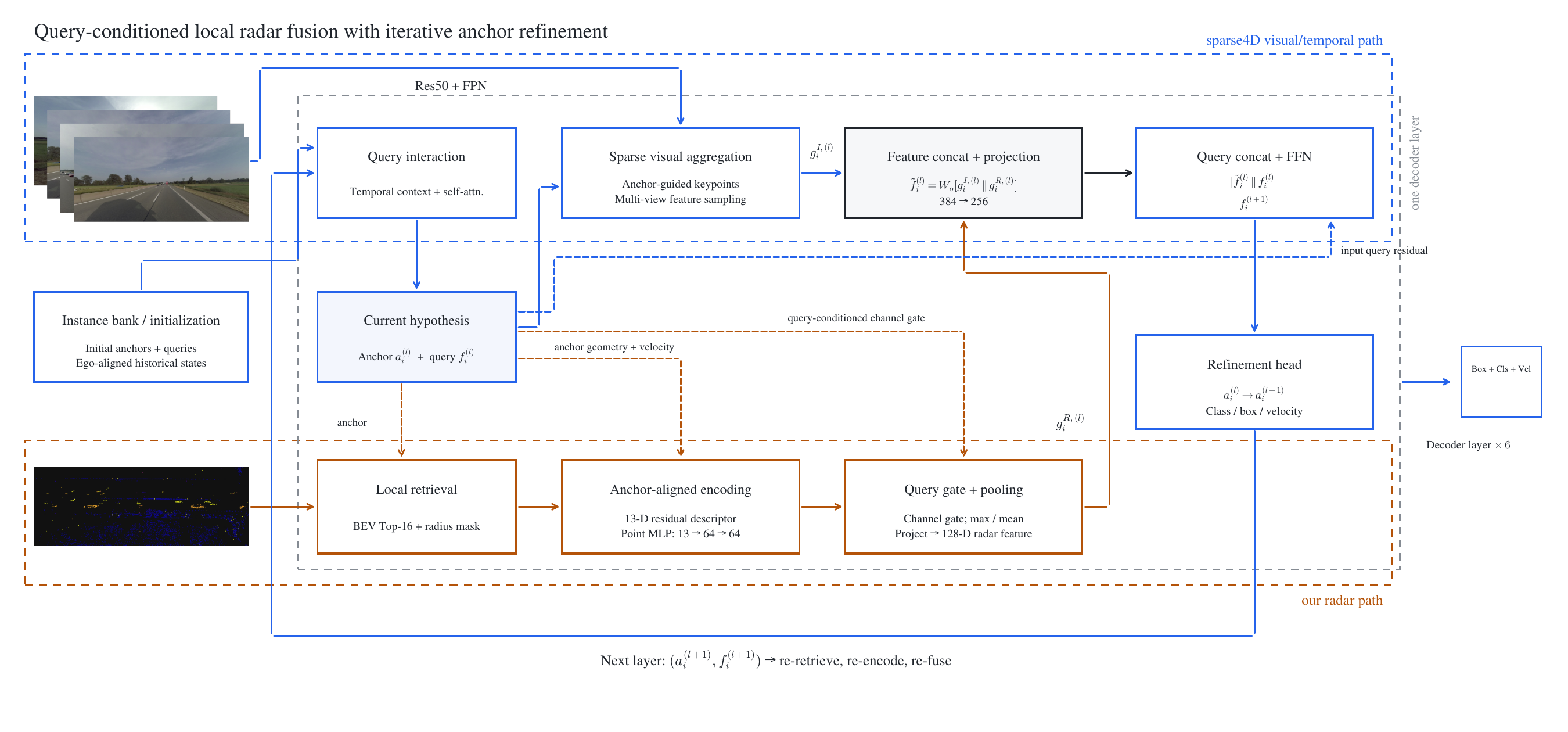}
\caption{Overview of \ours. Blue and orange dashed boxes denote the Sparse4D visual--temporal path and the local radar path, respectively; the gray dashed box outlines one decoder layer. Current anchors guide visual sampling and local radar retrieval, while query features modulate radar point encoding. Visual and radar features are concatenated and fused to update queries and anchors. Updated anchors trigger new radar retrieval at the next layer, forming an iterative loop. Six decoder layers produce box, class, and velocity predictions.}
\label{fig:overview}
\end{figure*}

This design casts radar fusion as evidence gathering for anchor refinement. It addresses three questions: how to retrieve hypothesis-relevant points with a fixed candidate budget, how to encode geometric and motion residuals relative to the current anchor, and how to condition the encoding on the query when candidate returns may originate from other objects. Accordingly, \ours{} performs anchor-conditioned local retrieval, anchor-aligned residual encoding, lightweight query-conditioned encoding, and iterative refinement. Fig.~\ref{fig:overview} illustrates the overall framework.

\noindent\textbf{Visual--temporal query backbone.}
We retain the sparse visual--temporal backbone of Sparse4Dv3 and modify only its multimodal aggregation path~\cite{sparse4dv3}. Multi-view images are encoded by ResNet-50 and a feature pyramid network~\cite{resnet,fpn}. For each query, 3D keypoints are generated from the current anchor and projected into the camera views. Deformable aggregation then samples multi-scale image features to obtain an object-centric visual representation.

The Sparse4Dv3 instance bank propagates historical queries. Historical anchors are transformed into the current reference frame using ego-motion, and their instance features provide temporal context for the decoder. Query self-attention models interactions between object hypotheses, and a refinement head updates anchors at each decoder layer. These visual and temporal operations are inherited from Sparse4Dv3; our changes are concentrated in the local raw radar path described below.

\subsection{Query-Conditioned Local Radar Encoding}
\noindent\textbf{Anchor-conditioned local retrieval.}
Current anchors provide an object-centric reference for selecting relevant radar points. For query $i$, let $o_i^{(l)}=[x_i^{(l)},y_i^{(l)}]$ denote the horizontal anchor center at layer $l$. We compute planar distances from this center to all radar points and select the nearest $K$ points as a fixed-budget candidate set. A size-dependent radius then filters these candidates:
\begin{equation}
\begin{aligned}
 d_{ij}^{(l)}&=\left\|o_i^{(l)}-p_j^{xy}\right\|_2,\\[-1pt]
 \mathcal{N}_i^{(l)}&=\operatorname{TopKIdx}_{K}\!\left(\{d_{ij}^{(l)}\}_{j=1}^{N}\right),\\[-1pt]
 m_{ij}^{(l)}&=\mathbb{I}\!\left[j\in\mathcal{N}_i^{(l)}\right]
 \mathbb{I}\!\left[d_{ij}^{(l)}\leq\rho_i^{(l)}\right].
\end{aligned}
\end{equation}
where $p_j^{xy}=[x_j,y_j]$, and $\operatorname{TopKIdx}_{K}$ returns the indices of the $K$ smallest distances. Let $w_i^{(l)}$ and $l_i^{(l)}$ denote the physical width and length of the anchor. The adaptive radius is
\begin{equation}
    \rho_i^{(l)}=\max\!\left(\rho_{\min},\alpha\left\|[w_i^{(l)},l_i^{(l)}]\right\|_2\right).
\end{equation}
We use $K=16$, $\rho_{\min}=2.0$~m, and $\alpha=0.75$. Top-$K$ retrieval yields a compact candidate set, while the adaptive radius determines geometric validity relative to the current hypothesis. Because retrieval uses the updated anchor at each layer, local radar evidence is repeatedly realigned as object estimates are refined.

\noindent\textbf{Residual descriptors for anchor refinement.}
Our objective is not to reconstruct the entire radar scene, but to use local evidence to refine the current anchor. Point descriptors should therefore express geometric and motion residuals relative to that hypothesis. Although the ego frame is convenient for data loading, it does not explicitly encode this relationship. We transform each valid point into the coordinate frame of the current anchor. Let $c_i=[x_i,y_i,z_i]$ be the 3D anchor center and $(w_i,l_i,h_i)$ its physical dimensions. Using the planar rotation matrix $R_2$, the relative position and velocity of point $j$ for query $i$ are
\begin{equation}
\begin{aligned}
 \Delta p_{ij}^{xy}&=p_j^{xy}-c_i^{xy},&
 \bar p_{ij}^{xy}&=R_2(-\theta_i)\Delta p_{ij}^{xy},\\[-1pt]
 \bar v_j&=R_2(-\theta_i)v_j,&
 \Delta\bar v_{ij}&=R_2(-\theta_i)(v_j-v_i).
\end{aligned}
\end{equation}
where $\theta_i$ and $v_i$ are the anchor yaw and horizontal velocity, respectively, and $\Delta z_{ij}=z_j-z_i$. Writing $\bar p_{ij}^{xy}=[\ell_{ij},a_{ij}]$, we normalize position by anchor dimensions as $\tilde p_{ij}=[\ell_{ij}/l_i,\,a_{ij}/w_i,\,\Delta z_{ij}/h_i]$. The resulting 13-dimensional descriptor is
\begin{equation}
    \begin{aligned}
    \phi_{ij}=\big[&
    \bar p_{ij}^{xy},\Delta z_{ij},\tilde p_{ij},
    \|\Delta p_{ij}^{xy}\|_2,\frac{\bar v_j}{10},\frac{\Delta\bar v_{ij}}{10},\\[-1pt]
    &\frac{s_j}{10},10\Delta t_j\big].
    \end{aligned}
\end{equation}
The first three channels describe position in the box frame. Size normalization provides a scale-aware representation, and the velocity residual expresses consistency between point motion and anchor motion. RCS and $\Delta t_j$ retain return strength and the temporal ordering of multi-sweep evidence. Motion-aware radar fusion has been explored in temporal radar--camera detectors and 4D radar encoders~\cite{crt_fusion,drnet,sgdet3d}; our focus is to organize these physical attributes into query-aligned descriptors for anchor refinement.

\noindent\textbf{Point encoding and query gating.}
Each query receives only a small, sparse candidate set. We therefore use a lightweight pointwise network rather than a dense radar backbone or cross-attention between points. Shared point encoding and symmetric set aggregation follow the basic principle of PointNet~\cite{pointnet}; we augment this principle with anchor-aligned geometric and motion residuals and query-conditioned channel modulation. Omitting layer superscripts below, a two-layer point MLP produces point features, while the current instance feature generates a channel-wise gate:
\begin{equation}
\begin{aligned}
 h_{ij}&=\operatorname{MLP}_{p}(\phi_{ij}),&
 g_i&=2\sigma(W_g f_i),\\[-1pt]
 \hat h_{ij}&=g_i\odot h_{ij}.&&
\end{aligned}
\end{equation}
The gate allows different object hypotheses to emphasize different radar attributes. It modulates feature channels without constructing costly point-to-point attention relationships. Conditioning on the current query also provides a mechanism for attenuating returns that are inconsistent with the hypothesis, such as those from nearby objects, the ground, or guardrails.

\noindent\textbf{Masked set aggregation.}
The number of valid radar points varies across objects, and some hypotheses receive no valid returns. To avoid treating padded or invalid entries as observations, we aggregate the $K$ candidate features in $\mathcal{N}_i^{(l)}$ using their validity masks. Layer superscripts are again omitted for clarity:
\begin{equation}
\begin{aligned}
 n_i&=\sum_jm_{ij},&
 \mu_i&=\frac{\sum_jm_{ij}\hat h_{ij}}{\max(n_i,1)},\\[-1pt]
 \mu_i^{\max}&=\max_{j:m_{ij}=1}\hat h_{ij},&
 z_i&=[\mu_i^{\max}\,\|\,\mu_i\,\|\,n_i/K].
\end{aligned}
\end{equation}
Mean pooling summarizes local context, max pooling retains the strongest channel responses, and $n_i/K$ explicitly represents the fraction of valid radar evidence available to the query. The masked maximum is evaluated only when $n_i>0$ and is set to zero otherwise. A projection maps the pooled representation to a 128-dimensional radar feature:
\begin{equation}
    g_i^{R}=\Phi_{p}(z_i)\in\mathbb{R}^{128}.
\end{equation}
When $n_i=0$, a learned no-radar embedding replaces the projected feature $g_i^R$. This explicitly distinguishes the absence of radar evidence from an observed radar feature whose numerical value happens to be zero.

\begin{figure*}[t]
\centering
\includegraphics[width=\textwidth]{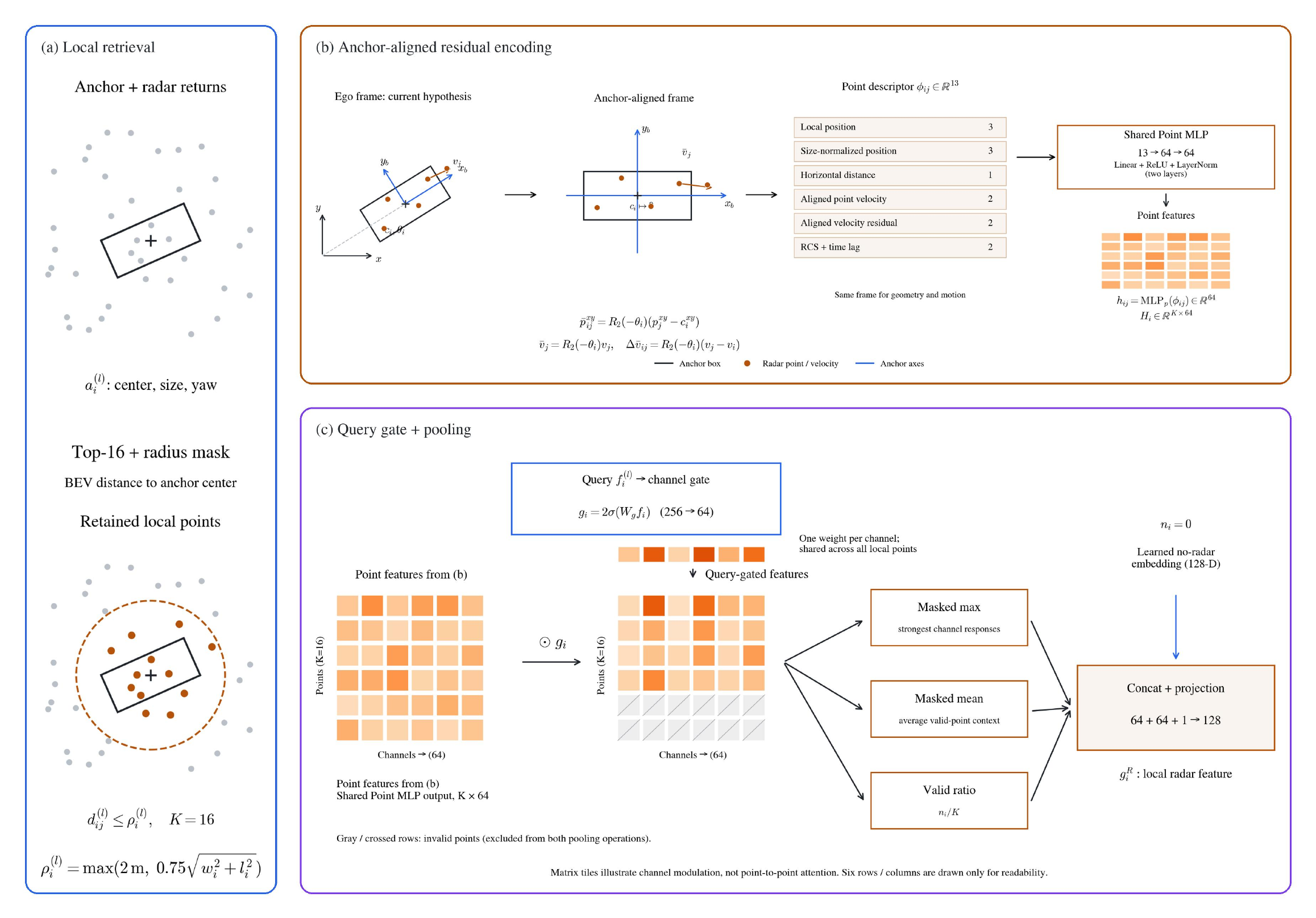}
\caption{Query-conditioned local radar branch. (a) Planar distance to the anchor center determines top-16 candidates, followed by radius filtering. (b) A 13-dimensional geometric and motion descriptor is constructed in the anchor frame and encoded by a shared point MLP. (c) Query-conditioned channel gating, masked max/mean pooling, and the valid-point ratio produce a local radar feature; empty neighborhoods use a learned embedding. Points and feature tiles are schematic illustrations.}
\label{fig:local_radar}
\end{figure*}

Fig.~\ref{fig:local_radar} summarizes the complete local radar path, from candidate retrieval and box-frame encoding to query gating and masked set aggregation.

\subsection{Iterative Cross-Modal Updates}
Local radar evidence contributes to anchor refinement together with visual sampling features. Let $g_i^{I,(l)}$ denote the Sparse4Dv3 visual aggregation feature and $g_i^{R,(l)}$ the 128-dimensional local radar feature. We concatenate and project them to the instance feature dimension, concatenate the result with the input query state, and apply a feed-forward network (FFN). A refinement head then predicts the next-layer anchor:
\begin{equation}
\begin{aligned}
 \tilde f_i^{(l)}&=W_o\left[g_i^{I,(l)}\,\|\,g_i^{R,(l)}\right],\\[-1pt]
 f_i^{(l+1)}&=\operatorname{FFN}\left(\left[\tilde f_i^{(l)}\,\|\,f_i^{(l)}\right]\right),\\[-1pt]
 a_i^{(l+1)}&=\operatorname{Refine}\left(f_i^{(l+1)},a_i^{(l)}\right).
\end{aligned}
\end{equation}
The updated anchor guides radar neighborhood retrieval at the next layer, forming a retrieval--encoding--fusion--refinement loop.
This loop couples radar evidence selection with object estimation across layers: changes in the anchor alter the next retrieval neighborhood, box coordinate frame, and velocity residuals, while the resulting radar evidence contributes to subsequent anchor refinement. Compared with a global Radar-BEV path, the design avoids dense radar feature extraction and concentrates computation around the current object hypotheses.

\subsection{Training and Implementation Details}
\noindent\textbf{Training objectives.} We adopt the detection and auxiliary objectives of the Sparse4Dv3-style backbone, including focal classification loss~\cite{focal}, 3D box and horizontal velocity regression, centerness and orientation quality losses, and auxiliary depth supervision from projected LiDAR points. LiDAR is not used at inference. These objectives follow the inherited visual--temporal training setup.

The local radar branch is trained jointly with the visual--temporal detector. Denoising queries and regular queries pass through the same local radar fusion path, exposing the encoder to both original and perturbed object hypotheses.

\noindent\textbf{Network configuration.}
Unless otherwise specified, we use a ResNet-50 backbone with FPN, 900 object queries, six decoder layers, and three radar sweeps from each of six sensors. Radar points have seven channels $(x,y,z,v_x,v_y,\mathrm{RCS},\Delta t)$, with at most 8192 points retained per sample and padding applied as needed. The local radar encoder uses a two-layer point MLP to map 13-dimensional descriptors to 64-dimensional point features, followed by set aggregation and projection to a 128-dimensional query-local radar feature. The local encoding branch contains approximately 0.4M parameters. Each query retrieves 16 nearest candidates, with a minimum radius of 2.0~m and radius scale of 0.75. Optimization settings and the training schedule are specified in the experiments.

\section{Experiments}
\subsection{Dataset and Evaluation Protocol}
We evaluate on MAN TruckScenes, which provides multi-view images, six 4D radar sensors, long-range 3D annotations, and truck-specific driving conditions~\cite{mantruckscenes}. We follow the official \texttt{detection\_cvpr\_2024} protocol with 12 object classes and center-distance matching thresholds of 0.5, 1, 2, and 4~m. The evaluation range is 150~m for the five vehicle classes and 75~m for the remaining classes. We report mAP, NDS, mATE, mASE, mAOE, mAVE, and mAAE on the official validation split. All reproduced systems use the same class mapping and evaluator.

\noindent\textbf{Training configuration.}
We use a ResNet-50 image backbone. The main configuration uses a per-camera network input resolution of 448$\times$960 (height$\times$width); an additional 256$\times$864 configuration examines the resolution trade-off and provides a size-matched comparison with published results. The model uses four cameras, three radar sweeps per sensor, 900 queries, and six decoder layers. Training uses AdamW with a learning rate of $3\times10^{-4}$, weight decay of $10^{-3}$, a global batch size of 24, and a 100-epoch cosine schedule with 500 warm-up iterations on two A800 GPUs. The inference efficiency measurements below use batch size 1 on an RTX 4090 D. Timing includes the model forward pass, temporal propagation, query-local radar search, and post-processing, but excludes PCD decoding, CPU preprocessing, and host-to-device transfer.

\subsection{Overall Detection Results}
We compare published methods with three reproduced systems under the same MAN TruckScenes evaluation protocol and class mapping. Sparse4Dv3 serves as the camera-only reference, while Sparse4D-Radar uses global Radar-BEV fusion. Table~\ref{tab:main_results} reports mAP, NDS, and the official true-positive errors. At 448$\times$960, QSRadar improves mAP by 21.22 points over the camera-only detector and exceeds the global Radar-BEV baseline by 1.01 mAP points and 1.42 NDS points. With the input reduced to 256$\times$864, QSRadar achieves 36.32 mAP and 50.16 NDS, exceeding the reported ResNet-50 results of Horizon3D at the same resolution.

\begin{table*}[t]
\caption{Accuracy on the official MAN TruckScenes validation split. mAP and NDS are percentages; the remaining metrics are official error measures. Published methods use the official 12-class evaluation protocol. $^{\dagger}$The 256$\times$864 EviScope result is from a single eight-epoch run without a matched baseline in the paper. $^{\ddagger}$The four-epoch EviScope and baseline results at 256$\times$704 are averages over seven runs.}
\label{tab:main_results}
\centering
\scriptsize
\setlength{\tabcolsep}{2.5pt}
\renewcommand{\arraystretch}{0.95}
\resizebox{\textwidth}{!}{%
\begin{tabular}{lccccccccccc}
\toprule
Method & Modality & Backbone & Image size & Epoch & NDS$\uparrow$ & mAP$\uparrow$ & mATE$\downarrow$ & mASE$\downarrow$ & mAOE$\downarrow$ & mAVE$\downarrow$ & mAAE$\downarrow$ \\
\midrule
RCTrans~\cite{rctrans} & C+R & R50 & 256$\times$864 & -- & 22.9 & 12.6 & 1.025 & 0.507 & 0.608 & 0.894 & 0.331 \\
CRT-Fusion~\cite{crt_fusion} & C+R & R50 & 256$\times$864 & -- & 28.8 & 16.9 & 0.833 & 0.512 & 0.444 & 0.852 & 0.322 \\
BEVFusion~\cite{bevfusion} & C+R & R50 & 256$\times$864 & -- & 30.4 & 18.2 & 0.941 & 0.442 & 0.389 & 0.892 & 0.208 \\
Horizon3D~\cite{horizon3d} & C+R & R50 & 256$\times$864 & 48 & 37.4 & 23.6 & 0.876 & 0.410 & 0.331 & 0.625 & 0.198 \\
SpaRC~\cite{sparc} & C+R & V2-99 & 640$\times$960 & -- & 35.4 & 22.5 & 0.798 & 0.449 & 0.476 & 0.613 & 0.248 \\
Horizon3D~\cite{horizon3d} & C+R & V2-99 & 640$\times$960 & 48 & 38.4 & 24.1 & 0.833 & 0.404 & 0.355 & 0.561 & 0.208 \\
EviScope baseline~\cite{eviscope}$^{\ddagger}$ & C+R & R50 & 256$\times$704 & 4 & 38.1 & 25.9 & 0.655 & 0.418 & 0.537 & 0.747 & 0.123 \\
EviScope~\cite{eviscope}$^{\ddagger}$ & C+R & R50 & 256$\times$704 & 4 & 38.4 & 25.6 & 0.650 & 0.418 & 0.519 & 0.722 & 0.126 \\
EviScope~\cite{eviscope} & C+R & R50 & 256$\times$704 & 8 & 39.9 & 26.7 & 0.654 & 0.408 & 0.480 & 0.671 & 0.129 \\
EviScope~\cite{eviscope}$^{\dagger}$ & C+R & R50 & 256$\times$864 & 8 & 39.9 & 28.2 & 0.599 & 0.412 & 0.568 & 0.719 & 0.127 \\
\midrule
Sparse4Dv3 (reproduced) & C & R50 & 448$\times$960 & 100 & 26.78 & 16.87 & 0.7189 & 0.4901 & 0.6258 & 1.0061 & 0.3308 \\
Sparse4D-Radar (global BEV) & C+R & R50 & 448$\times$960 & 60 & 50.78 & 37.08 & 0.4542 & 0.3955 & 0.4224 & 0.4294 & 0.0751 \\
\ours (256$\times$864) & C+R & R50 & 256$\times$864 & 100 & 50.16 & 36.32 & 0.5222 & 0.4017 & 0.4373 & 0.3668 & 0.0722 \\
\ours (448$\times$960) & C+R & R50 & 448$\times$960 & 80 & \textbf{52.20} & \textbf{38.09} & 0.4621 & 0.3885 & 0.3804 & 0.3769 & 0.0769 \\
\bottomrule
\end{tabular}%
}
\end{table*}

Fig.~\ref{fig:sota_fps_map} compares our 256$\times$864 measurements with publicly reported MAN TruckScenes results. It includes BEVFusion, CRT-Fusion, RCTrans, and Horizon3D results compiled under the official validation protocol by Horizon3D~\cite{horizon3d}. QSRadar lies in the high-accuracy, high-throughput region among the displayed methods. Because the hardware and timing scopes differ, the plot provides efficiency context rather than a strictly controlled speed ranking. Methods such as SpaRC, which report official accuracy but not throughput, remain in Table~\ref{tab:main_results} but are omitted from this plot.

\subsection{Error and Range Analysis}
The true-positive errors in Table~\ref{tab:main_results} show that QSRadar achieves lower scale, orientation, and velocity errors among the reproduced systems, while Sparse4D-Radar has slightly lower translation and attribute errors. These results indicate that the compact query-local radar path retains geometric and motion cues relevant to object estimation.

\noindent\textbf{Range-dependent results.}
Table~\ref{tab:range_results} further examines performance across distance bands using the official matching and filtering rules. QSRadar improves Car AP over the global Radar-BEV baseline in every band, including 100--150~m. For Trailer beyond 25~m, however, the global Radar-BEV model remains stronger and also achieves higher overall band-wise mAP at the farthest range. QSRadar therefore offers a favorable long-range trade-off for the prevalent Car class while remaining competitive on truck-related categories.

\begin{table}[tb]
\caption{Range-wise diagnostic results on MAN TruckScenes. Matching, class-specific range limits, and ground-truth filtering follow the official evaluator. $\mathrm{mAP}_{\mathrm{present}}$ averages only over classes with ground-truth instances in each distance band.}
\label{tab:range_results}
\centering
\scriptsize
\begin{tabular}{lcccc}
\toprule
Method & Range & mAP$_{\mathrm{present}}$ & Car AP & Trailer AP\\
\midrule
Sparse4Dv3 & 0--25 m & 0.3940 & 0.7458 & 0.7151\\
Sparse4D-Radar & 0--25 m & 0.4913 & 0.8674 & 0.7671\\
\ours & 0--25 m & 0.5013 & 0.8849 & 0.7600\\
Sparse4Dv3 & 25--50 m & 0.1590 & 0.3223 & 0.2675\\
Sparse4D-Radar & 25--50 m & 0.3675 & 0.7175 & 0.5803\\
\ours & 25--50 m & 0.3857 & 0.7386 & 0.5062\\
Sparse4Dv3 & 50--100 m & 0.0498 & 0.0676 & 0.0761\\
Sparse4D-Radar & 50--100 m & 0.2907 & 0.6564 & 0.4436\\
\ours & 50--100 m & 0.3093 & 0.6855 & 0.3705\\
Sparse4Dv3 & 100--150 m & 0.0110 & 0.0020 & 0.0425\\
Sparse4D-Radar & 100--150 m & 0.3088 & 0.5727 & 0.5101\\
\ours & 100--150 m & 0.2824 & 0.5885 & 0.4048\\
\bottomrule
\end{tabular}
\end{table}

\subsection{Efficiency Analysis}
The matched RTX 4090 D measurements in Table~\ref{tab:efficiency_4090d} use batch size 1, 448$\times$960 inputs, temporal propagation, and the same forward-pass and post-processing scope. QSRadar achieves 19.24 FPS with 0.798~GB peak memory usage and 48.59M parameters. Relative to global Radar-BEV fusion, its throughput is 1.83$\times$ higher, while peak memory usage and GFLOPs are reduced to 51\% and 57\% of the baseline, respectively.

\begin{table}[tb]
\caption{Matched inference efficiency on an RTX 4090 D. All methods use batch size 1, 448$\times$960 inputs, temporal propagation, and the same forward-pass and post-processing timing scope.}
\label{tab:efficiency_4090d}
\centering
\scriptsize
\setlength{\tabcolsep}{4pt}
\begin{tabular}{lcccc}
\toprule
Method & FPS & Peak memory (GB) & Params. (M) & GFLOPs\\
\midrule
Sparse4Dv3 & \textbf{23.55} & \textbf{0.796} & \textbf{48.16} & \textbf{533.4}\\
Sparse4D-Radar & 10.50 & 1.569 & 53.72 & 933.1\\
\ours & 19.24 & 0.798 & 48.59 & 535.4\\
\bottomrule
\end{tabular}
\end{table}

\subsection{Inference Sensitivity and Qualitative Analysis}
We perform inference-time sensitivity tests on a single trained QSRadar checkpoint, measuring the effects of changing radar input representations, neighborhood retrieval, or set aggregation paths. Results are reported in Table~\ref{tab:design_ablation}. None of the variants is retrained. These tests therefore characterize the trained model's dependence on its existing inputs and representations; they do not establish the relative merits of alternative architectures trained independently.

\begin{table}[tb]
\caption{Inference-time sensitivity of the local radar path using the same checkpoint and official evaluator, without retraining. ``Reuse after layer 2'' retains separate retrieval in layers 1 and 2 and reuses layer 2 neighborhoods in layers 3--6; 0.39~ms is the retrieval time for this setting.}
\label{tab:design_ablation}
\centering
\scriptsize
\setlength{\tabcolsep}{4.5pt}
\begin{tabular}{llccc}
\toprule
Study & Variant & mAP$\uparrow$ & NDS$\uparrow$ & Key metric\\
\midrule
\multirow{3}{*}{Retrieval}
& Full ($K{=}16$) & \textbf{0.3798} & \textbf{0.5247} & --\\
& Fixed radius & 0.3672 & 0.5243 & --\\
& Reuse after layer 2 & 0.3773 & 0.5175 & Search: 0.39 ms\\
\midrule
\multirow{4}{*}{Descriptor}
& Full box frame & \textbf{0.3798} & \textbf{0.5247} & --\\
& Ego frame & 0.2347 & 0.3565 & --\\
& No velocity residual & 0.2825 & 0.3844 & mAVE: 1.8645\\
& No RCS + $\Delta t$ & 0.3045 & 0.4355 & --\\
\midrule
\multirow{4}{*}{Aggregation}
& Full gate + mean/max & \textbf{0.3798} & 0.5247 & --\\
& No query gate & 0.3366 & 0.4926 & Car: 0.6249\\
& Mean pooling only & 0.2367 & 0.3823 & Car: 0.4970\\
& Max pooling only & 0.3267 & 0.4553 & Car: 0.6733\\
\bottomrule
\end{tabular}
\end{table}

For this trained checkpoint, changing the box-frame representation, removing velocity residuals, or altering the pooling path reduces performance, indicating dependence on these inputs and representations. A fixed retrieval radius or neighborhood reuse in later layers also changes detection performance. Removing box-frame rotation and retaining only mean pooling lead to large mAP drops, while removing velocity residuals increases mAVE to 1.8645, indicating sensitivity of this model's velocity estimates to that input. These observations neither isolate the causal contribution of individual components nor imply that alternative designs would necessarily perform worse after retraining.

\noindent\textbf{Qualitative results.}
\begin{figure*}[t]
\centering
\includegraphics[width=\textwidth]{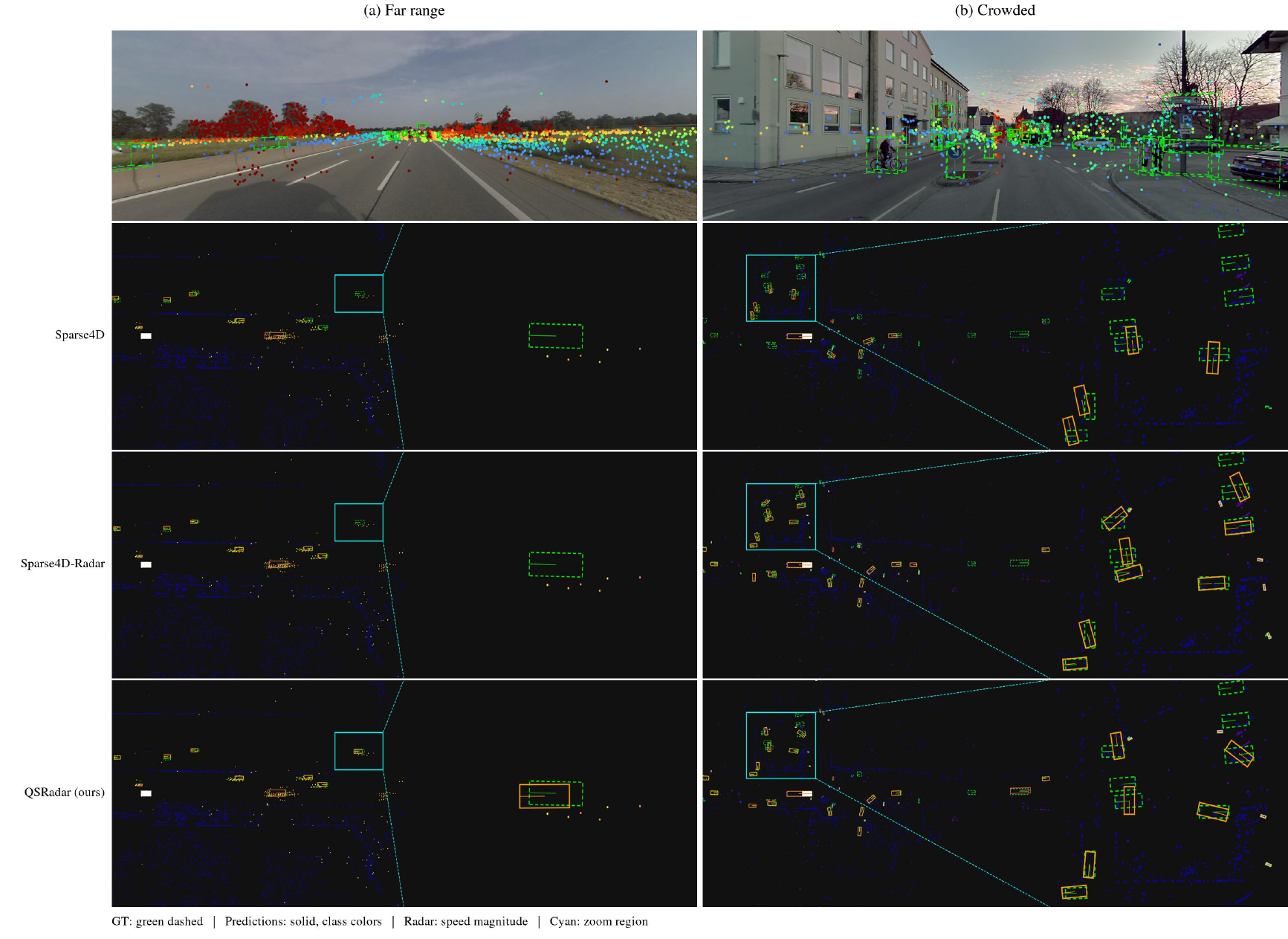}
\caption{Qualitative detection results in long-range and crowded scenes. Shared camera views are shown at the top, followed by BEV results from Sparse4D, Sparse4D-Radar, and QSRadar. Green dashed boxes denote ground truth, solid colored boxes denote predictions, and BEV radar points are colored by speed magnitude. Cyan rectangles and dashed lines indicate the regions enlarged on the right. All methods use identical display ranges and a confidence threshold of 0.3.}
\label{fig:detection_results}
\end{figure*}

Fig.~\ref{fig:detection_results} compares QSRadar with camera-only Sparse4D and the global Radar-BEV fusion baseline Sparse4D-Radar in long-range and crowded scenes. All methods use identical scene extents, zoom regions, and confidence thresholds. Camera views provide appearance and visibility context, while the BEV views and enlargements show object coverage and box localization.

In the long-range scene in Fig.~\ref{fig:detection_results}(a), the highlighted vehicle is approximately 126~m from the ego vehicle and occupies only a small image region, making its position and dimensions difficult to estimate from appearance alone. At the display threshold, neither camera-only Sparse4D nor the global fusion baseline produces a corresponding prediction at this vehicle, whereas QSRadar predicts a vehicle box near the ground truth. This example illustrates the potential of query-conditioned local radar fusion for long-range detection. It does not, however, identify the input attribute or internal mechanism responsible for the improvement, and the inference sensitivity tests do not directly quantify the independent contribution of each component.

In the crowded scene in Fig.~\ref{fig:detection_results}(b), nearby objects occupy a confined area, some are partially occluded, and radar observations may include returns from multiple objects. The enlarged region contains several vehicles, allowing comparison of object coverage and box localization. Both radar fusion methods produce more vehicle predictions in this region than the camera-only model. QSRadar retains the ability to detect multiple neighboring vehicles without constructing a global Radar-BEV feature map. This observation is consistent with our design objective: concentrating radar computation around current object hypotheses while retaining useful radar-assisted detection capability. These visualizations illustrate selected examples; overall accuracy and efficiency are assessed by the quantitative experiments.

\section{Conclusion}
We presented \ours{}, an efficient query-conditioned sparse radar--camera fusion framework for 3D object detection. Built on a Sparse4Dv3-style visual--temporal detector, \ours{} retrieves local raw radar evidence around evolving object hypotheses, encodes it in hypothesis-aligned coordinates, and aggregates it through query-conditioned set pooling. The design avoids dense Radar-BEV construction while retaining radar geometry, motion, signal strength, and timing information. On the official MAN TruckScenes validation set, the main checkpoint achieves 38.09 mAP and 52.20 NDS, compared with 37.08 mAP and 50.78 NDS for the reproduced Sparse4D-Radar global Radar-BEV baseline. Matched inference tests on an RTX 4090 D show reductions of 45\% in latency, 49\% in peak memory usage, and 43\% in GFLOPs, yielding a favorable accuracy--efficiency trade-off.


\end{document}